\pdfoutput=1
\documentclass[11pt]{article}

\usepackage[final]{acl}

\usepackage{times}
\usepackage{latexsym}
\usepackage{amsmath}
\usepackage{dblfloatfix}
\usepackage[T1]{fontenc}
\usepackage{array,multirow,graphicx}
\usepackage{rotating}
\usepackage{diagbox}
\usepackage{caption}
\usepackage{tabularx}
\usepackage{calc}
\usepackage[normalem]{ulem}
\usepackage[utf8]{inputenc}

\usepackage{microtype}

\usepackage{inconsolata}

\title{Effectiveness of Chain-of-Thought in Distilling Reasoning Capability from Large Language Models}

\author{Cong-Thanh Do$^{1,}$\thanks{\textbf{Correspondence:} \href{mailto:cong-thanh.do@toshiba.eu}{cong-thanh.do@toshiba.eu}}, Rama Doddipatla$^{1}$, and Kate Knill$^{2}$ \\
      $^{1}$Cambridge Research Laboratory, Toshiba Europe Ltd., Cambridge, UK \\ $^{2}$Department of Engineering, University of Cambridge, Cambridge, UK}

\begin{document}

\newcommand{\STAB}[1]{\begin{tabular}{@{}c@{}}#1\end{tabular}}

\maketitle
\begin{abstract}
Chain-of-Thought (CoT) prompting is a widely used method to improve the reasoning capability of Large Language Models (LLMs). More recently, CoT has been leveraged in Knowledge Distillation (KD) to transfer reasoning capability from a larger LLM to a smaller one. This paper examines the role of CoT in distilling the reasoning capability from larger LLMs to smaller LLMs using white-box KD, analysing its effectiveness in improving the performance of the distilled models for various natural language reasoning and understanding tasks. We conduct white-box KD experiments using LLMs from the Qwen and Llama2 families, employing CoT data from the CoT-Collection dataset. The distilled models are then evaluated on natural language reasoning and understanding tasks from the BIG-Bench-Hard (BBH) benchmark, which presents complex challenges for smaller LLMs. Experimental results demonstrate the role of CoT in improving white-box KD effectiveness, enabling the distilled models to achieve better average performance in natural language reasoning and understanding tasks from BBH.
\end{abstract}

\section{Introduction}

 \begin{figure*}[h]
	\centering
  \includegraphics[width=1.9\columnwidth]{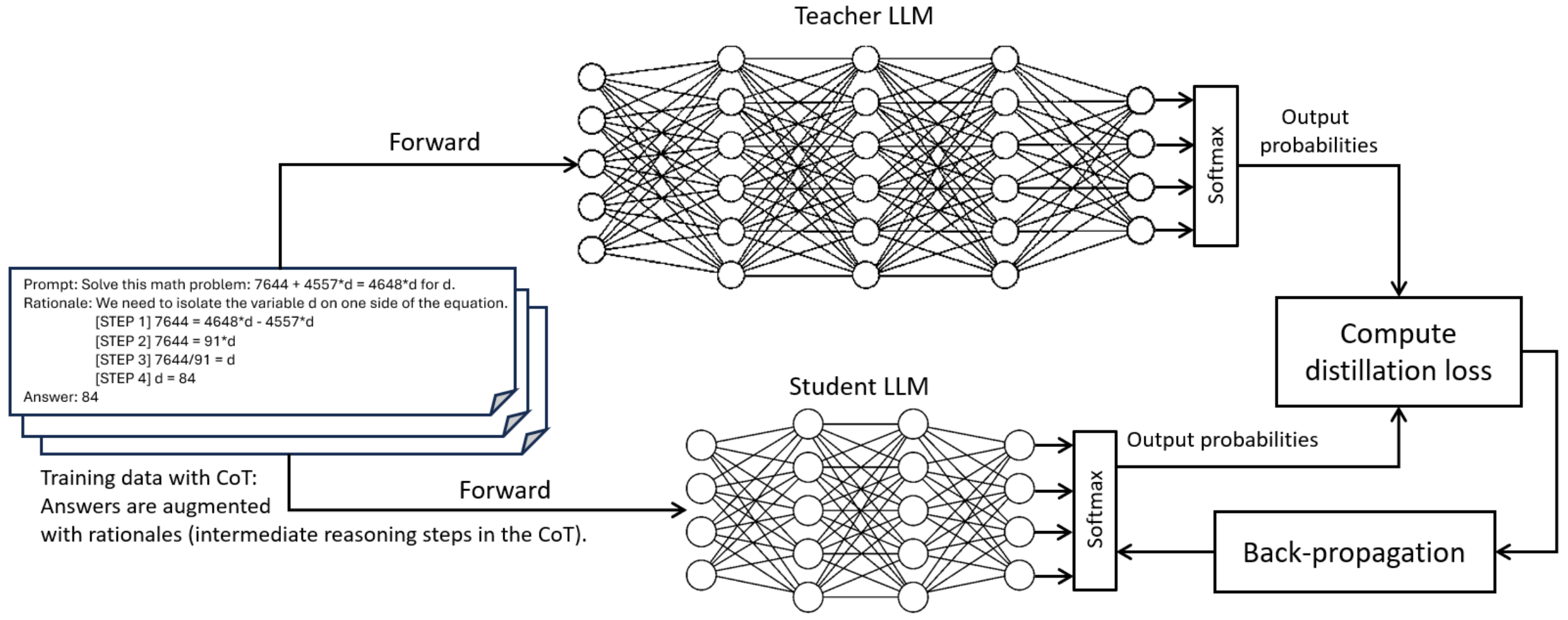}
  \caption{KD+CoT approach for distilling reasoning capability from a teacher LLM to a student LLM using white-box KD and CoT. Rationales—intermediate reasoning steps in CoT— are incorporated in the training data of KD+CoT. These rationales are not incorporated in the vanilla white-box KD's training data.\vspace{-5pt}}
  \label{fig:knowledge_distillation}
\end{figure*}

Reasoning is the process of using logic, evidence, and knowledge to make sense of information, draw conclusions, solve problems, and make decisions. While reasoning is considered as one of the essential capabilities, it tends to emerge only in Large Language Models (LLMs) at a certain scale \cite{wei2022_tmlr}. Smaller LLMs, typically with fewer than tens of billions of parameters, have limited reasoning capability \cite{fu2023specializingsmallerlanguagemodels}. Improving the reasoning capability of LLMs, especially of smaller ones, is an important research topic with various practical applications \cite{lu2024smalllanguagemodelssurvey}.% including LLM-based dialogue systems. 

Indeed, improving the reasoning capability for smaller LLMs enables real-time responsiveness, delivering faster processing than larger models—an essential factor for interactive applications utilising natural language. These more efficient LLMs facilitate edge deployment on devices with limited resources, addressing the computational demands of larger models. Furthermore, better reasoning helps smaller models understand nuanced language and maintain context throughout extended natural language interactions, even in the presence of linguistic imperfections \cite{beygi-etal-2022-logical, dongre-etal-2025-respact}. Finally, they offer reduced operational cost for inference, making them more accessible, and empower natural language processing systems to be more proactive in problem-solving and task completion.

Several approaches have been developed to improve the reasoning capabilities of LLMs. For example, techniques like few-shot prompting \cite{kojima2022_neurips} and Supervised Fine-Tuning (SFT) \cite{devlin-etal-2019-bert} can improve the capabilities of LLMs, including reasoning, in specific domains when training data are available. Chain-of-Thought (CoT) prompting is a popular method used to elicit the reasoning capabilities of LLMs \cite{wei2022_neurips}. CoT, which involves a series of intermediate reasoning steps, significantly improves the capability of LLMs to perform complex reasoning tasks \cite{huang2023_acl_findings, ling2023_neurips}.

Knowledge Distillation (KD) \cite{hinton2014} is a promising technique to transfer knowledge, including reasoning capability, from larger LLMs to smaller LLMs. This technique is frequently employed to condense the knowledge stored in larger neural networks into smaller counterparts, thereby reducing computational resource requirements and improving inference speed without substantial performance sacrifices. More recently, CoT has been leveraged in KD to transfer multi-step reasoning capability from a larger LLM to a smaller one \cite{wang-etal-2023-scott, lee2024mentorkdmakingsmalllanguage}. In \cite{lee2024mentorkdmakingsmalllanguage}, a mentor model, which is an intermediate-sized, task-specific fine-tuned model, was used to augment additional CoT annotations and provide soft labels for the student model during reasoning distillation.

While CoT enhances complex reasoning in LLMs, integrating it with white-box KD—where the teacher model's internal workings are fully visible during distillation—presents opportunities and challenges that require deeper investigation. In this paper, we examine the role of CoT in the distillation of reasoning capability from larger LLMs to smaller LLMs using white-box KD. We demonstrate the effectiveness of CoT in improving KD performance, enabling the distilled models to achieve better average performance in various natural language reasoning and understanding tasks from the BIG-Bench-Hard benchmark \cite{suzgun2023_acl_findings}.

\vspace{-5pt}
\section{Related work}
\label{sec:related_work}

CoT distillation is a popular approach which is used to distill reasoning capability from larger LLMs to smaller LLMs \cite{magister2023_acl_short, shridhar2023_acl_findings, ho2023_acl_long, wang-etal-2023-scott, chen2023_emnlp_findings, li2024_lrec_coling, feng2024_icml}. The main idea in these works is to prompt a very large teacher LLM, such as GPT-4 \cite{openai2024gpt4technicalreport}, to solve complex questions via zero-shot CoT prompting \cite{wei2022_neurips}, and then use the reasoning samples to fine-tune a smaller student LLM. This approach is applied with black-box KD \cite{chen2024knowledgedistillationblackboxlarge}, where only the texts generated by the teacher LLMs are accessible.

White-box KD approach \cite{hinton2014} leverages full access to the teacher LLM's output probabilities or intermediate hidden states during distillation. This approach enables the student LLM to learn not just the final outputs but the underlying reasoning process—leading to improved performance on complex language understanding and reasoning tasks \cite{zhang2024_emnlp}. With the availability of more open-source LLMs, white-box KD is becoming increasingly valuable for both academia and industry, as it provides good potential to improve performance. However, most studies on white-box KD have focused on traditional small language models \cite{gu2024_iclr}, and the application of CoT in white-box KD has yet to be widely explored.

\section{KD+CoT: White-box Knowledge Distillation with Chain-of-Thought}
\label{sec:cot_kd}

In this paper, we examine the role of CoT in the distillation of reasoning capability from larger LLMs to smaller LLMs using white-box KD, where the output probabilities of the teacher LLM can be accessed during the distillation process \cite{hinton2014}. We assume that the teacher and student LLMs have tokenizers of the same size $N$ and the same vocabulary, to simplify the computation of the distillation loss between probability distributions. We use Kullback-Leibler (KL) divergence as the only loss in the distillation loss to keep the KD step pure with only information flowing from the teacher model \cite{sreenivas2024llmpruningdistillationpractice}.

The KD+CoT approach for distilling reasoning capability from the teacher LLM to the student LLM is illustrated in Fig. \ref{fig:knowledge_distillation}. During the KD+CoT process, training instances with rationales are forwarded through both the teacher and student LLMs. The output probabilities produced at the output layers of these models are used to compute the distillation loss. The error derivative of the distillation loss is then back-propagated through the student LLM to update its weights and minimise the distillation loss. The key difference between KD+CoT and vanilla white-box KD is that the former incorporates rationales in the training data, while the latter does not.

\begin{figure}[t]
  \includegraphics[width=1\columnwidth]{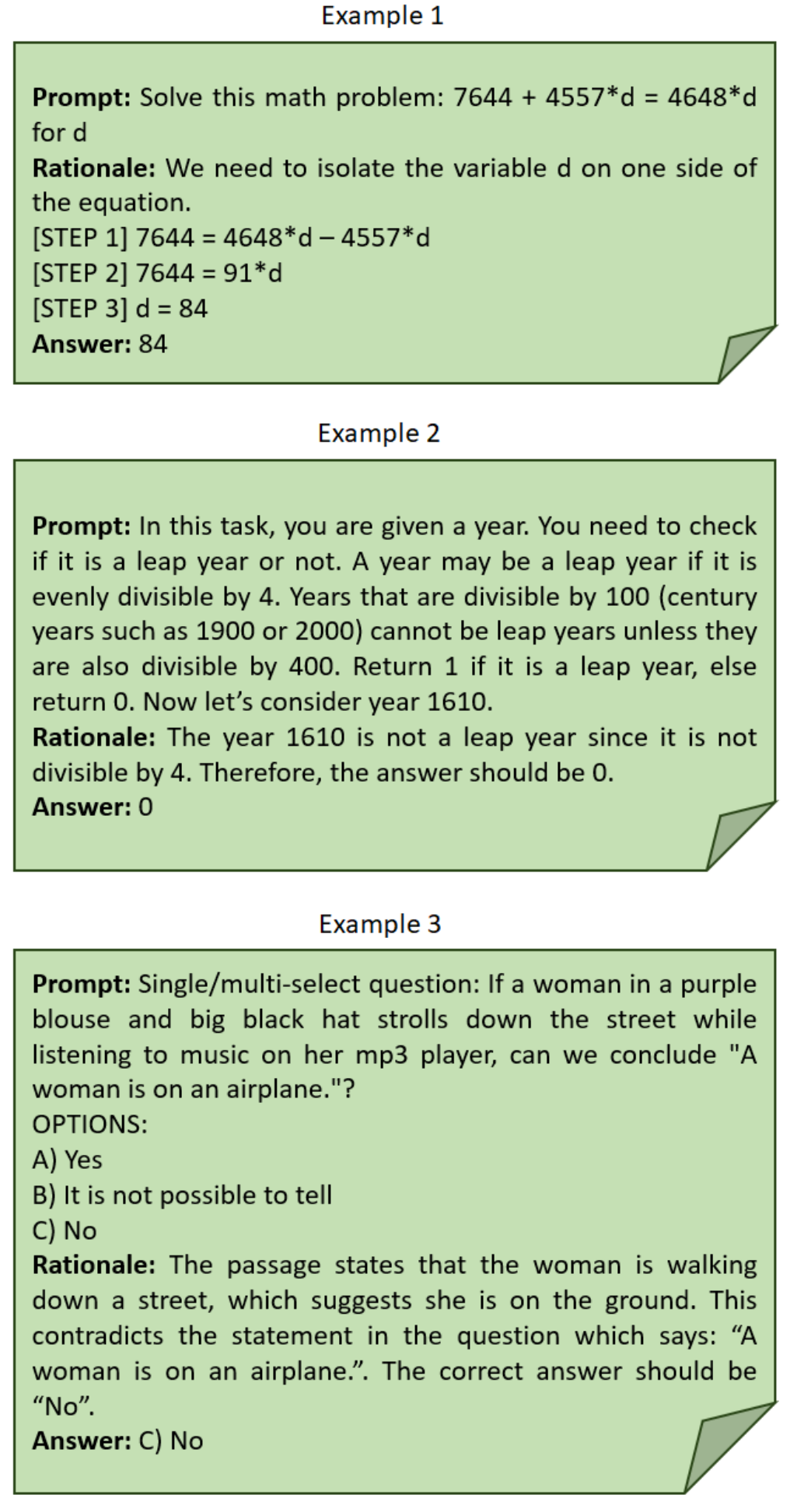}
  \caption{Examples of training data instances with integrated rationales that lead to the answers. These data are from the CoT-Collection dataset \cite{kim2023_emnlp}.\vspace{-15pt}}
  \label{fig:cot}
\end{figure}

Figure \ref{fig:cot} illustrates examples of training instances from the CoT-Collection dataset \cite{kim2023_emnlp}, showcasing integrated step-by-step reasoning rationales. Incorporating these rationales into the training data for KD is expected to elicit the teacher LLM's reasoning capability during training, effectively distilling it to the student LLM by minimising the distillation loss between the models' output probabilities. 

\vspace{-5pt}
\section{Experiments}
\label{sec:experiments}
\vspace{-5pt}
\subsection{Models, data, and training}

\subsubsection{Models}
\label{sec:models}

We use two families of LLMs for the experiments, Qwen \cite{bai2023qwentechnicalreport} and Llama2 \cite{touvron2023llama2openfoundation}. These models are chosen for their proficiency in few-shot learning and their unique vocabulary coverage. The Qwen-1.8B and Qwen-7B models which have 1.8 and 7 billion parameters, respectively, were proposed by Alibaba Cloud. These models are used as the student and teacher LLMs, respectively, in the Qwen-based KD experiments. Qwen-1.8B and Qwen-7B are Transformer-based LLMs, which were pre-trained on a large volume of data, including web texts, books, codes, etc. The Qwen tokenizer consists of $N=151,936$ tokens. 

In Llama2-based KD experiments, the Llama2-13B-Chat is used as the teacher model, and the Llama2-7B and TinyLlama are used as the student models. Llama2 is an auto-regressive language model that employs an optimised Transformer architecture  and was trained on 2 trillions tokens of data. The Llama2-13B-Chat model, which consists of 13 billion parameters, is a fine-tuned version of the Llama2-13B model. It utilises SFT and reinforcement learning with human feedback (RLHF) to align with human preferences for helpfulness and safety. The Llama2-7B model has 7 billion parameters, while the TinyLlama model \cite{zhang2024tinyllamaopensourcesmalllanguage} has 1.1 billion parameters and was pre-trained on 1 trillion tokens of data. TinyLlama is built on the architecture and tokenizer of Llama2 which consists of $N=32,000$ tokens.

\subsubsection{Data}
\label{sec:data}

A CoT is a series of intermediate natural language reasoning steps that lead to the final output. These intermediate steps, also known as rationales, are crucial for understanding the reasoning process. We use the CoT-Collection dataset \cite{kim2023_emnlp}, an instruction tuning dataset that includes 1.84 million rationales from the FLAN collection \cite{longpre2023_icml} across 1,060 tasks, such as \textit{multiple choices QA}, \textit{extractive QA}, \textit{closed-book QA}, \textit{formal logic}, \textit{natural language inference}, and \textit{arithmetic}. These rationales were augmented using OpenAI Codex \cite{openaicodex}. We separate 1.84 million instances of the CoT-Collection into training and validation sets, with the training set consisting of 1.44 million instances and the development set consisting of 0.4 million instances.

\subsubsection{Training}
\label{sec:kd_and_sft}
%\vspace{-10pt}

We implement our training recipes based on the MiniLLM framework \cite{gu2024_iclr}. The KD is trained over 20,000 training steps during 10 epochs, saving intermediate models every 1,000 steps. All intermediate models resulting from a KD training are evaluated during inference, and the best score is selected and reported in section \ref{sec:results_short}. In our training, 20,000 iterations are sufficient to ensure the KD is sufficiently trained, given the training and validation data from the CoT-Collection dataset. We integrate the Low-Rank Adaptation (LoRA) algorithm \cite{hu2022_iclr} into the KD process to reduce computational requirements when Llama2-7B is used as student model. LoRA freezes the pre-trained model weights and injects trainable rank decomposition matrices into each layer of the Transformer architecture, significantly reducing the number of trainable parameters and, consequently, the computational demands during KD. In our LoRA setting, we set the rank $r = 32$, $alpha = 32$, and the drop-out rate is set to 0.1. A KD training completes in about 30 hours using an NVIDIA A100 80GB GPU card. A learning rate of 5e-6 and a batch size of 16 are used. The temperature parameter $\tau$ is set to 1 and the max prompt length of the model is set to 512.

\subsection{Evaluation}
\label{sec:inference_details}

The BBH benchmark \cite{suzgun2023_acl_findings} is used in this study. BBH is a benchmark comprising 23 challenging natural language reasoning and understanding tasks from BIG-Bench \cite{srivastava2023_tmlr}, specifically those where prior language model evaluations did not surpass human performance. In total, BBH includes 27 tasks, as some tasks have sub-tasks. The 27 tasks in BBH can be regrouped into four categories as follows: \textbf{algorithmic and multi-step arithmetic reasoning} (e.g. \textit{multistep arithmetic two}, \textit{boolean expressions}, \textit{logical deduction (three objects, five objects}, and \textit{seven objects}), \textit{geometric shapes}, \textit{dyck languages}, \textit{navigate}, and \textit{temporal sequences}), \textbf{natural language understanding} (e.g., \textit{disambiguation QA}, \textit{hyperbaton} (adjective ordering), and \textit{snarks} (sarcasm detection)), \textbf{use of world knowledge} (e.g., \textit{sport understanding}, \textit{movie recommendation}, \textit{date understanding}, \textit{causal judgement}, and \textit{ruin names}), and \textbf{multilingual knowledge and reasoning} (e.g., \textit{salient translation error detection}).% These tasks are unrelated to the training data from the CoT-Collection dataset \cite{kim2023_emnlp} which were used for the KD and SFT.

To enhance the performance of all models on BBH, few-shot prompting with CoT \cite{wei2022_neurips} is utilised during inference and is independent with training. This involves providing the models with a few demonstrations of the task at inference time as conditioning \cite{brown2020_neurips}. Specifically, we use the same few-shot prompting with CoT as in \cite{suzgun2023_acl_findings}, which includes three demonstrations. During inference, the temperature parameter is set to 0.2 and the batch size is set to 16.

\subsubsection{Results}
\label{sec:results_short}

\begin{table*}[t]
  \centering
  \begin{tabular}{|l|l|l|l|l|c|}
    \hline
    \fontsize{8}{9}\selectfont Task  & \fontsize{8}{9}\selectfont \# Questions & \fontsize{8}{9}\selectfont Qwen-1.8B & \fontsize{8}{9}\selectfont Qwen-1.8B+KD & \fontsize{8}{9}\selectfont Qwen-1.8B+KD+CoT & \fontsize{8}{9}\selectfont Teacher \\
    \hline
    \fontsize{9}{9}\selectfont Boolean expressions  & \fontsize{9}{9}\selectfont 3013 & \fontsize{9}{9}\selectfont 43.20  & \fontsize{9}{9}\selectfont 42.40 (\textcolor{red}{-1.85\%}) & \fontsize{9}{9}\selectfont 34.80 (\textcolor{red}{-19.44\%}) & \fontsize{9}{9}\selectfont 76.40  \\
    \fontsize{9}{9}\selectfont Dyck languages  & \fontsize{9}{9}\selectfont 10729 & \fontsize{9}{9}\selectfont 0.00  & \fontsize{9}{9}\selectfont 0.00 (\textcolor{black}{0.00\%}) & \fontsize{9}{9}\selectfont 0.00 (\textcolor{black}{0.00\%}) & \fontsize{9}{9}\selectfont 17.20  \\
		\fontsize{9}{9}\selectfont Formal fallacies  & \fontsize{9}{9}\selectfont 23062 & \fontsize{9}{9}\selectfont 11.60  & \fontsize{9}{9}\selectfont 29.60 (\textcolor{blue}{+155.17\%}) & \fontsize{9}{9}\selectfont 36.40 (\textcolor{blue}{+213.79\%}) & \fontsize{9}{9}\selectfont 49.60  \\
		\fontsize{9}{9}\selectfont Geometric shapes  & \fontsize{9}{9}\selectfont 8416 & \fontsize{9}{9}\selectfont 1.20 & \fontsize{9}{9}\selectfont 0.80 (\textcolor{red}{-33.33\%}) & \fontsize{9}{9}\selectfont 0.00 (\textcolor{red}{-100.00\%}) & \fontsize{9}{9}\selectfont 18.40  \\						
\fontsize{9}{9}\selectfont Logical deduction (3 objects)  & \fontsize{9}{9}\selectfont 17561 & \fontsize{9}{9}\selectfont 25.60 & \fontsize{9}{9}\selectfont 30.00 (\textcolor{blue}{+17.19\%}) & \fontsize{9}{9}\selectfont 32.80 (\textcolor{blue}{+28.13\%}) & \fontsize{9}{9}\selectfont 54.00  \\
		\fontsize{9}{9}\selectfont Logical deduction (5 objects)  & \fontsize{9}{9}\selectfont 24932 & \fontsize{9}{9}\selectfont 11.60  & \fontsize{9}{9}\selectfont 16.40 (\textcolor{blue}{+41.38\%}) & \fontsize{9}{9}\selectfont 18.00 (\textcolor{blue}{+55.17\%}) & \fontsize{9}{9}\selectfont 36.80 \\
		\fontsize{9}{9}\selectfont Logical deduction (7 objects)  & \fontsize{9}{9}\selectfont 32094 & \fontsize{9}{9}\selectfont 4.40  & \fontsize{9}{9}\selectfont 11.60 (\textcolor{blue}{+163.64\%}) & \fontsize{9}{9}\selectfont 8.80 (\textcolor{blue}{+100.00\%}) & \fontsize{9}{9}\selectfont 32.40 \\
\fontsize{9}{9}\selectfont Multistep arithmetic two  & \fontsize{9}{9}\selectfont 4763 & \fontsize{9}{9}\selectfont 8.40 & \fontsize{9}{9}\selectfont 0.00 (\textcolor{red}{-100.00\%)} & \fontsize{9}{9}\selectfont 2.00 (\textcolor{red}{-76.19\%)} & \fontsize{9}{9}\selectfont 26.40  \\
		\fontsize{9}{9}\selectfont Navigate  & \fontsize{9}{9}\selectfont 8759 & \fontsize{9}{9}\selectfont 52.00 & \fontsize{9}{9}\selectfont 40.00 (\textcolor{red}{-23.08\%}) & \fontsize{9}{9}\selectfont 40.80 (\textcolor{red}{-21.54\%})  & \fontsize{9}{9}\selectfont 58.80 \\
		\fontsize{9}{9}\selectfont Temporal sequences  & \fontsize{9}{9}\selectfont 23912 & \fontsize{9}{9}\selectfont 14.80 & \fontsize{9}{9}\selectfont 18.00 (\textcolor{blue}{+21.62\%}) & \fontsize{9}{9}\selectfont 28.40 (\textcolor{blue}{+91.89\%}) & \fontsize{9}{9}\selectfont 38.80  \\
    \fontsize{9}{9}\selectfont Tracking shuffled objects (3 objects)  & \fontsize{9}{9}\selectfont 20824 & \fontsize{9}{9}\selectfont 10.80 & \fontsize{9}{9}\selectfont 32.80 (\textcolor{blue}{+203.70\%}) & \fontsize{9}{9}\selectfont 32.00 (\textcolor{blue}{+196.30\%}) & \fontsize{9}{9}\selectfont 33.20  \\		
		\fontsize{9}{9}\selectfont Tracking shuffled objects (5 objects)  & \fontsize{9}{9}\selectfont 27607 & \fontsize{9}{9}\selectfont 2.80 & \fontsize{9}{9}\selectfont 13.60 (\textcolor{blue}{+385.71\%}) & \fontsize{9}{9}\selectfont 15.20 (\textcolor{blue}{+442.86\%}) & \fontsize{9}{9}\selectfont 21.60 \\
		\fontsize{9}{9}\selectfont Tracking shuffled objects (7 objects)  & \fontsize{9}{9}\selectfont 34328 & \fontsize{9}{9}\selectfont 0.00 & \fontsize{9}{9}\selectfont 6.00 (\textcolor{blue}{+6.00\%, abs.}) & \fontsize{9}{9}\selectfont 11.60 (\textcolor{blue}{11.60\%, abs.}) & \fontsize{9}{9}\selectfont 13.60  \\
		\fontsize{9}{9}\selectfont Word sorting  & \fontsize{9}{9}\selectfont 7775 & \fontsize{9}{9}\selectfont 1.60 & \fontsize{9}{9}\selectfont 1.20 (\textcolor{red}{-25.00\%}) & \fontsize{9}{9}\selectfont 1.60 (\textcolor{black}{0.00\%}) & \fontsize{9}{9}\selectfont 17.20  \\								
		\hline
    \fontsize{9}{9}\selectfont Disambiguation QA  & \fontsize{9}{9}\selectfont 12049 & \fontsize{9}{9}\selectfont 39.20  & \fontsize{9}{9}\selectfont 33.60 (\textcolor{red}{-14.29\%}) & \fontsize{9}{9}\selectfont 30.00 (\textcolor{red}{-23.47\%})  & \fontsize{9}{9}\selectfont 43.20  \\
		\fontsize{9}{9}\selectfont Hyperbaton  & \fontsize{9}{9}\selectfont 5135 & \fontsize{9}{9}\selectfont 9.20 & \fontsize{9}{9}\selectfont 48.40 (\textcolor{blue}{+426.08\%}) & \fontsize{9}{9}\selectfont 51.20 (\textcolor{blue}{+456.52\%}) & \fontsize{9}{9}\selectfont 78.40  \\
		\fontsize{9}{9}\selectfont Snarks  & \fontsize{9}{9}\selectfont 6206 & \fontsize{9}{9}\selectfont 38.20 & \fontsize{9}{9}\selectfont 43.26 (\textcolor{blue}{+13.25\%}) & \fontsize{9}{9}\selectfont 43.26 (\textcolor{blue}{+13.25\%}) & \fontsize{9}{9}\selectfont 59.55 \\
		\hline
    \fontsize{9}{9}\selectfont Causal judgement  & \fontsize{9}{9}\selectfont 35034 & \fontsize{9}{9}\selectfont 2.14  & \fontsize{9}{9}\selectfont 3.20 (\textcolor{blue}{+49.53\%}) & \fontsize{9}{9}\selectfont 9.10 (\textcolor{blue}{+325.23\%}) & \fontsize{9}{9}\selectfont 59.36   \\
    \fontsize{9}{9}\selectfont Date understanding  & \fontsize{9}{9}\selectfont 7530 & \fontsize{9}{9}\selectfont 26.40  & \fontsize{9}{9}\selectfont 31.60 (\textcolor{blue}{+19.70\%}) & \fontsize{9}{9}\selectfont 37.60 (\textcolor{blue}{+42.42\%}) & \fontsize{9}{9}\selectfont 61.20  \\
		\fontsize{9}{9}\selectfont Movie recommendation  & \fontsize{9}{9}\selectfont 7881 & \fontsize{9}{9}\selectfont 34.40  & \fontsize{9}{9}\selectfont 24.80 (\textcolor{red}{-27.91\%}) & \fontsize{9}{9}\selectfont 27.60 (\textcolor{red}{-19.77\%}) & \fontsize{9}{9}\selectfont 71.20 \\
		\fontsize{9}{9}\selectfont Object counting  & \fontsize{9}{9}\selectfont 6425 & \fontsize{9}{9}\selectfont 26.40 & \fontsize{9}{9}\selectfont 22.80 (\textcolor{red}{-13.64\%}) & \fontsize{9}{9}\selectfont 28.40 (\textcolor{blue}{+7.58\%}) & \fontsize{9}{9}\selectfont 68.00  \\
		\fontsize{9}{9}\selectfont Penguins in a table  & \fontsize{9}{9}\selectfont 13396 & \fontsize{9}{9}\selectfont 7.53 & \fontsize{9}{9}\selectfont 17.12 (\textcolor{blue}{+127.36\%}) & \fontsize{9}{9}\selectfont 21.92 (\textcolor{blue}{+191.10\%}) & \fontsize{9}{9}\selectfont 52.74  \\										
		\fontsize{9}{9}\selectfont Reasoning about colored objects  & \fontsize{9}{9}\selectfont 13554 & \fontsize{9}{9}\selectfont 2.40  & \fontsize{9}{9}\selectfont 16.40 (\textcolor{blue}{+583.33\%}) & \fontsize{9}{9}\selectfont 14.40 (\textcolor{blue}{+500.00\%}) & \fontsize{9}{9}\selectfont 62.00 \\
		\fontsize{9}{9}\selectfont Ruin names  & \fontsize{9}{9}\selectfont 7617 & \fontsize{9}{9}\selectfont 16.80 & \fontsize{9}{9}\selectfont 20.00 (\textcolor{blue}{+19.05\%}) & \fontsize{9}{9}\selectfont 22.40 (\textcolor{blue}{+33.33\%})  & \fontsize{9}{9}\selectfont 42.40  \\		
		\fontsize{9}{9}\selectfont Sport understanding  & \fontsize{9}{9}\selectfont 3780 & \fontsize{9}{9}\selectfont 32.40 & \fontsize{9}{9}\selectfont 60.80 (\textcolor{blue}{+87.65\%}) & \fontsize{9}{9}\selectfont 56.40 (\textcolor{blue}{+74.07\%}) & \fontsize{9}{9}\selectfont 75.20  \\
		\fontsize{9}{9}\selectfont Web of lies  & \fontsize{9}{9}\selectfont 8035 & \fontsize{9}{9}\selectfont 47.60  & \fontsize{9}{9}\selectfont 49.60 (\textcolor{blue}{+4.20\%}) & \fontsize{9}{9}\selectfont 48.80 (\textcolor{blue}{+2.52\%}) & \fontsize{9}{9}\selectfont 68.80 \\																																				
		\hline
		\fontsize{9}{9}\selectfont Salient translation error detection  & \fontsize{9}{9}\selectfont 38692 & \fontsize{9}{9}\selectfont 9.20  & \fontsize{9}{9}\selectfont 9.60 (\textcolor{blue}{+4.35\%}) & \fontsize{9}{9}\selectfont 6.40 (\textcolor{red}{-30.43\%}) & \fontsize{9}{9}\selectfont 42.80 \\		
    \hline
		\fontsize{9}{9}\selectfont Average  & \fontsize{9}{9}\selectfont 15300 & \fontsize{9}{9}\selectfont 17.77\% & \fontsize{9}{9}\selectfont 23.10\% (\textcolor{blue}{+30.00\%}) & \fontsize{9}{9}\selectfont \textbf{24.44\%} (\textbf{\textcolor{blue}{+37.54\%}}) & \fontsize{9}{9}\selectfont 47.38\% \\
		\hline
  \end{tabular}
  \caption{\label{performance-qwen-tasks}
   Exact-match scores (in \%) of the student (baseline), distilled (KD), and teacher models on the 27 tasks of the BBH benchmark, with Qwen-1.8B as the student model and Qwen-7B as the teacher model. The numbers in parentheses represent relative gains, except where absolute gains (abs.) are specified. The number (\#) of questions for each task is also displayed. The tasks are grouped into four categories in accordance with the order presented in section \ref{app:discussion}. \vspace{-10pt}
  }
\end{table*}

\begin{table*}[t]
  \centering
  \begin{tabular}{|l|l|l|l|l|c|}
    \hline
    \fontsize{8}{8}\selectfont Task  & \fontsize{8}{8}\selectfont \# Questions & \fontsize{8}{8}\selectfont Llama2-7B & \fontsize{8}{8}\selectfont Llama2-7B+KD & \fontsize{8}{8}\selectfont Llama2-7B+KD+CoT & \fontsize{8}{8}\selectfont Teacher  \\
    \hline
    \fontsize{9}{9}\selectfont Boolean expressions  & \fontsize{9}{9}\selectfont 3013 & \fontsize{9}{9}\selectfont 68.00  & \fontsize{9}{9}\selectfont 70.40 (\textcolor{blue}{+3.52\%}) & \fontsize{9}{9}\selectfont 72.00 (\textcolor{blue}{+5.88\%}) & \fontsize{9}{9}\selectfont 72.40 \\
    \fontsize{9}{9}\selectfont Dyck languages  & \fontsize{9}{9}\selectfont 10729 & \fontsize{9}{9}\selectfont 8.40  & \fontsize{9}{9}\selectfont 8.00 (\textcolor{red}{-4.76\%}) & \fontsize{9}{9}\selectfont 12.00 (\textcolor{blue}{+42.85\%}) & \fontsize{9}{9}\selectfont 32.40 \\
		\fontsize{9}{9}\selectfont Formal fallacies  & \fontsize{9}{9}\selectfont 23062 & \fontsize{9}{9}\selectfont 49.60  & \fontsize{9}{9}\selectfont 50.40 (\textcolor{blue}{+1.61\%}) & \fontsize{9}{9}\selectfont 51.20 (\textcolor{blue}{+3.22\%}) & \fontsize{9}{9}\selectfont 49.60  \\
		\fontsize{9}{9}\selectfont Geometric shapes  & \fontsize{9}{9}\selectfont 8416 & \fontsize{9}{9}\selectfont 30.00  & \fontsize{9}{9}\selectfont 32.00 (\textcolor{blue}{+6.66\%}) & \fontsize{9}{9}\selectfont 32.00 (\textcolor{blue}{+6.66\%})  & \fontsize{9}{9}\selectfont 37.20 \\						
		\fontsize{9}{9}\selectfont Logical deduction (3 objects)  & \fontsize{9}{9}\selectfont 17561 & \fontsize{9}{9}\selectfont 56.40 & \fontsize{9}{9}\selectfont 53.60 (\textcolor{red}{-4.96\%}) & \fontsize{9}{9}\selectfont 56.80 (\textcolor{blue}{+0.70\%}) & \fontsize{9}{9}\selectfont 73.60  \\
		\fontsize{9}{9}\selectfont Logical deduction (5 objects)  & \fontsize{9}{9}\selectfont 24932 & \fontsize{9}{9}\selectfont 33.60 & \fontsize{9}{9}\selectfont 31.60 (\textcolor{red}{-5.95\%}) & \fontsize{9}{9}\selectfont 33.60 (0\%) & \fontsize{9}{9}\selectfont 50.40  \\
		\fontsize{9}{9}\selectfont Logical deduction (7 objects)  & \fontsize{9}{9}\selectfont 32094 & \fontsize{9}{9}\selectfont 22.00  & \fontsize{9}{9}\selectfont 26.80 (\textcolor{blue}{+21.81\%}) & \fontsize{9}{9}\selectfont 30.40 (\textcolor{blue}{+38.18\%}) & \fontsize{9}{9}\selectfont 40.80 \\
		\fontsize{9}{9}\selectfont Multistep arithmetic two  & \fontsize{9}{9}\selectfont 4763 & \fontsize{9}{9}\selectfont 0.00 & \fontsize{9}{9}\selectfont 0.40 (\textcolor{blue}{+0.40\%, abs.)} & \fontsize{9}{9}\selectfont 2.00 (\textcolor{blue}{+2.00\%, abs.)} & \fontsize{9}{9}\selectfont 4.40 \\
		\fontsize{9}{9}\selectfont Navigate  & \fontsize{9}{9}\selectfont 8759 & \fontsize{9}{9}\selectfont 54.40 & \fontsize{9}{9}\selectfont 58.40 (\textcolor{blue}{+7.35\%}) & \fontsize{9}{9}\selectfont 59.60 (\textcolor{blue}{+9.55\%}) & \fontsize{9}{9}\selectfont 64.00 \\
		\fontsize{9}{9}\selectfont Temporal sequences  & \fontsize{9}{9}\selectfont 23912 & \fontsize{9}{9}\selectfont 12.00 & \fontsize{9}{9}\selectfont 14.80 (\textcolor{blue}{+23.33\%}) & \fontsize{9}{9}\selectfont 12.80 (\textcolor{blue}{+6.66\%}) & \fontsize{9}{9}\selectfont 20.80 \\
    \fontsize{9}{9}\selectfont Tracking shuffled objects (3 objects)  & \fontsize{9}{9}\selectfont 20824 & \fontsize{9}{9}\selectfont 33.60 & \fontsize{9}{9}\selectfont 33.60 (\textcolor{black}{0\%}) & \fontsize{9}{9}\selectfont 32.00 (\textcolor{red}{-4.76\%}) & \fontsize{9}{9}\selectfont 34.40  \\		
		\fontsize{9}{9}\selectfont Tracking shuffled objects (5 objects)  & \fontsize{9}{9}\selectfont 27607 & \fontsize{9}{9}\selectfont 15.60 & \fontsize{9}{9}\selectfont 18.00 (\textcolor{blue}{+15.38\%}) & \fontsize{9}{9}\selectfont 16.00 (\textcolor{blue}{+2.56\%}) & \fontsize{9}{9}\selectfont 21.20  \\
		\fontsize{9}{9}\selectfont Tracking shuffled objects (7 objects)  & \fontsize{9}{9}\selectfont 34328 & \fontsize{9}{9}\selectfont 11.20  & \fontsize{9}{9}\selectfont 17.20 (\textcolor{blue}{+53.57\%}) & \fontsize{9}{9}\selectfont 14.40 (\textcolor{blue}{+28.57\%}) & \fontsize{9}{9}\selectfont 18.40 \\
		\fontsize{9}{9}\selectfont Word sorting  & \fontsize{9}{9}\selectfont 7775 & \fontsize{9}{9}\selectfont 10.40 & \fontsize{9}{9}\selectfont 5.60 (\textcolor{red}{-46.15\%}) & \fontsize{9}{9}\selectfont 4.80 (\textcolor{red}{-53.84\%}) & \fontsize{9}{9}\selectfont 24.40  \\
\hline												
    \fontsize{9}{9}\selectfont Disambiguation QA  &\fontsize{9}{9}\selectfont 12049 & \fontsize{9}{9}\selectfont 46.80  & \fontsize{9}{9}\selectfont 50.80 (\textcolor{blue}{+8.54\%}) & \fontsize{9}{9}\selectfont 50.40 (\textcolor{blue}{+7.69\%}) & \fontsize{9}{9}\selectfont 47.20  \\
		\fontsize{9}{9}\selectfont Hyperbaton  & \fontsize{9}{9}\selectfont 5135 & \fontsize{9}{9}\selectfont 52.40  & \fontsize{9}{9}\selectfont 54.00 (\textcolor{blue}{+3.05\%}) & \fontsize{9}{9}\selectfont 60.80 (\textcolor{blue}{+16.03\%}) & \fontsize{9}{9}\selectfont 57.60 \\
		\fontsize{9}{9}\selectfont Snarks  & \fontsize{9}{9}\selectfont 6206 & \fontsize{9}{9}\selectfont 47.75 & \fontsize{9}{9}\selectfont 51.12 (\textcolor{blue}{+7.05\%}) & \fontsize{9}{9}\selectfont 57.30 (\textcolor{blue}{+20.00\%}) & \fontsize{9}{9}\selectfont 70.78 \\
\hline
    \fontsize{9}{9}\selectfont Causal judgement  & \fontsize{9}{9}\selectfont 35034 & \fontsize{9}{9}\selectfont 52.40  & \fontsize{9}{9}\selectfont 50.26 (\textcolor{red}{-4.08\%}) & \fontsize{9}{9}\selectfont 49.19 (\textcolor{red}{-6.12\%}) & \fontsize{9}{9}\selectfont 59.89 \\
    \fontsize{9}{9}\selectfont Date understanding  & \fontsize{9}{9}\selectfont 7530 & \fontsize{9}{9}\selectfont 60.80  & \fontsize{9}{9}\selectfont 63.60 (\textcolor{blue}{+4.60\%}) & \fontsize{9}{9}\selectfont 62.40 (\textcolor{blue}{+2.63\%}) & \fontsize{9}{9}\selectfont 68.00  \\
		\fontsize{9}{9}\selectfont Movie recommendation  & \fontsize{9}{9}\selectfont 7881 & \fontsize{9}{9}\selectfont 68.00 & \fontsize{9}{9}\selectfont 53.60 (\textcolor{red}{-21.17\%}) & \fontsize{9}{9}\selectfont 67.20 (\textcolor{red}{-1.17\%}) & \fontsize{9}{9}\selectfont 64.80  \\				
		\fontsize{9}{9}\selectfont Object counting  & \fontsize{9}{9}\selectfont 6425 & \fontsize{9}{9}\selectfont 50.00  & \fontsize{9}{9}\selectfont 49.60 (\textcolor{red}{-0.8\%}) & \fontsize{9}{9}\selectfont 49.20 (\textcolor{red}{-1.60\%})  & \fontsize{9}{9}\selectfont 55.60 \\
		\fontsize{9}{9}\selectfont Penguins in a table  & \fontsize{9}{9}\selectfont 13396 & \fontsize{9}{9}\selectfont 32.87  & \fontsize{9}{9}\selectfont 33.56 (\textcolor{blue}{+2.09\%}) & \fontsize{9}{9}\selectfont 35.61 (\textcolor{blue}{+8.33\%})  & \fontsize{9}{9}\selectfont 51.36 \\
		\fontsize{9}{9}\selectfont Reasoning about colored objects  & \fontsize{9}{9}\selectfont 13554 & \fontsize{9}{9}\selectfont 48.00 & \fontsize{9}{9}\selectfont 45.60 (\textcolor{red}{-5.00\%}) & \fontsize{9}{9}\selectfont 49.60 (\textcolor{blue}{+3.33\%}) & \fontsize{9}{9}\selectfont 62.40  \\
		\fontsize{9}{9}\selectfont Ruin names  & \fontsize{9}{9}\selectfont 7617 & \fontsize{9}{9}\selectfont 33.20 & \fontsize{9}{9}\selectfont 28.80 (\textcolor{red}{-13.25\%}) & \fontsize{9}{9}\selectfont 29.60 (\textcolor{red}{-10.84\%}) & \fontsize{9}{9}\selectfont 41.20  \\
		\fontsize{9}{9}\selectfont Sport understanding  & \fontsize{9}{9}\selectfont 3780 & \fontsize{9}{9}\selectfont 88.00 & \fontsize{9}{9}\selectfont 84.80 (\textcolor{red}{-3.63\%}) & \fontsize{9}{9}\selectfont 88.40 (\textcolor{blue}{+0.45\%}) & \fontsize{9}{9}\selectfont 92.00 \\								
		\fontsize{9}{9}\selectfont Web of lies  & \fontsize{9}{9}\selectfont 8035 & \fontsize{9}{9}\selectfont 54.40 & \fontsize{9}{9}\selectfont 49.60 (\textcolor{red}{-8.82\%}) & \fontsize{9}{9}\selectfont 61.20 (\textcolor{blue}{+12.5\%}) & \fontsize{9}{9}\selectfont 94.40 \\								
\hline		
		\fontsize{9}{9}\selectfont Salient translation error detection  & \fontsize{9}{9}\selectfont 38692 & \fontsize{9}{9}\selectfont 25.20 & \fontsize{9}{9}\selectfont 22.80 (\textcolor{red}{-9.52\%}) & \fontsize{9}{9}\selectfont 30.00 (\textcolor{blue}{+19.04\%}) & \fontsize{9}{9}\selectfont 39.60 \\
    \hline
		\fontsize{9}{9}\selectfont Average  & \fontsize{9}{9}\selectfont 15300 & \fontsize{9}{9}\selectfont 39.44\% & \fontsize{9}{9}\selectfont 39.22\% (\textcolor{red}{-0.56\%}) & \fontsize{9}{9}\selectfont \textbf{41.50\%} (\textbf{\textcolor{blue}{+5.22\%}}) & \fontsize{9}{9}\selectfont 49.95\% \\
		\hline
  \end{tabular}
  \caption{\label{performance-llama2-tasks}
   Exact-match scores (in \%) of the models on 27 tasks of BBH, with Llama2-7B as the student model and Llama2-13B-Chat as the teacher model. The numbers in parentheses represent relative gains, except where absolute gains (abs.) are specified. The Llama2-7B's performance here is comparable to that reported in \cite{wan2024_iclr}.\vspace{-5pt}
  }
\end{table*}

%\addtolength{\tabcolsep}{-0.2em}
\begin{table*}[t]
  \centering
  \begin{tabular}{|l|l|l|l|l|c|}
    \hline
    \fontsize{8}{8}\selectfont Task  & \fontsize{8}{8}\selectfont \# Questions & \fontsize{8}{8}\selectfont TinyLlama & \fontsize{8}{8}\selectfont TinyLlama+KD & \fontsize{8}{8}\selectfont TinyLlama+KD+CoT & \fontsize{8}{8}\selectfont Teacher \\
    \hline
    \fontsize{9}{9}\selectfont Boolean expressions  & \fontsize{9}{9}\selectfont 3013 & \fontsize{9}{9}\selectfont 60.80  & \fontsize{9}{9}\selectfont 56.80 (\textcolor{red}{-6.57\%}) & \fontsize{9}{9}\selectfont 44.80 (\textcolor{red}{-26.31\%}) & \fontsize{9}{9}\selectfont 72.40 \\
    \fontsize{9}{9}\selectfont Dyck languages  & \fontsize{9}{9}\selectfont 10729 & \fontsize{9}{9}\selectfont 12.80  & \fontsize{9}{9}\selectfont 32.80 (\textcolor{blue}{+156.25\%}) & \fontsize{9}{9}\selectfont 27.60 (\textcolor{blue}{+115.62\%}) & \fontsize{9}{9}\selectfont 32.40 \\
		\fontsize{9}{9}\selectfont Formal fallacies  & \fontsize{9}{9}\selectfont 23062 & \fontsize{9}{9}\selectfont 47.60  & \fontsize{9}{9}\selectfont 36.00 (\textcolor{red}{-24.36\%}) & \fontsize{9}{9}\selectfont 49.20 (\textcolor{blue}{+3.36\%}) & \fontsize{9}{9}\selectfont 49.60  \\
		\fontsize{9}{9}\selectfont Geometric shapes  & \fontsize{9}{9}\selectfont 8416 & \fontsize{9}{9}\selectfont 0.00 & \fontsize{9}{9}\selectfont 0.00 (\textcolor{black}{0\%}) & \fontsize{9}{9}\selectfont 0.00 (\textcolor{black}{0\%}) & \fontsize{9}{9}\selectfont 37.20 \\						
		\fontsize{9}{9}\selectfont Logical deduction (3 objects)  & \fontsize{9}{9}\selectfont 17561 & \fontsize{9}{9}\selectfont 29.20 & \fontsize{9}{9}\selectfont 31.60 (\textcolor{red}{-26.02\%}) & \fontsize{9}{9}\selectfont 33.60 (\textcolor{blue}{+15.06\%}) & \fontsize{9}{9}\selectfont 73.60  \\		
		\fontsize{9}{9}\selectfont Logical deduction (5 objects)  & \fontsize{9}{9}\selectfont 24932 & \fontsize{9}{9}\selectfont 18.80 & \fontsize{9}{9}\selectfont 17.20 (\textcolor{red}{-8.51\%}) & \fontsize{9}{9}\selectfont 19.60 (\textcolor{blue}{+4.25\%}) & \fontsize{9}{9}\selectfont 50.40 \\
		\fontsize{9}{9}\selectfont Logical deduction (7 objects)  & \fontsize{9}{9}\selectfont 32094 & \fontsize{9}{9}\selectfont 12.40 & \fontsize{9}{9}\selectfont 10.00 (\textcolor{red}{-19.35\%}) & \fontsize{9}{9}\selectfont 14.80 (\textcolor{blue}{+19.35\%}) & \fontsize{9}{9}\selectfont 40.80  \\
		\fontsize{9}{9}\selectfont Multistep arithmetic two  & \fontsize{9}{9}\selectfont 4763 & \fontsize{9}{9}\selectfont 0.00 & \fontsize{9}{9}\selectfont 0.40 (\textcolor{blue}{+0.40\%, abs.}) & \fontsize{9}{9}\selectfont 0.80 (\textcolor{blue}{+0.80\%, abs.}) & \fontsize{9}{9}\selectfont 4.40 \\
		\fontsize{9}{9}\selectfont Navigate  & \fontsize{9}{9}\selectfont 8759 & \fontsize{9}{9}\selectfont 47.20 & \fontsize{9}{9}\selectfont 44.40 (\textcolor{red}{-5.93\%}) & \fontsize{9}{9}\selectfont 52.40 (\textcolor{blue}{+11.01\%}) & \fontsize{9}{9}\selectfont 64.00  \\
		\fontsize{9}{9}\selectfont Temporal sequences  & \fontsize{9}{9}\selectfont 23912 & \fontsize{9}{9}\selectfont 27.60 & \fontsize{9}{9}\selectfont 29.20 (\textcolor{blue}{+5.79\%}) & \fontsize{9}{9}\selectfont 25.60 (\textcolor{red}{-7.24\%}) & \fontsize{9}{9}\selectfont 20.80 \\						
		\fontsize{9}{9}\selectfont Tracking shuffled objects (3 objects)  & \fontsize{9}{9}\selectfont 20824 & \fontsize{9}{9}\selectfont 30.00 & \fontsize{9}{9}\selectfont 34.00 (\textcolor{blue}{+13.33\%}) & \fontsize{9}{9}\selectfont 30.80 (\textcolor{blue}{+2.66\%}) & \fontsize{9}{9}\selectfont 34.40 \\		
		\fontsize{9}{9}\selectfont Tracking shuffled objects (5 objects)  & \fontsize{9}{9}\selectfont 27607 & \fontsize{9}{9}\selectfont 17.60 & \fontsize{9}{9}\selectfont 19.20 (\textcolor{blue}{+9.09\%}) & \fontsize{9}{9}\selectfont 18.00 (\textcolor{blue}{+2.27\%}) & \fontsize{9}{9}\selectfont 21.20  \\
		\fontsize{9}{9}\selectfont Tracking shuffled objects (7 objects)  & \fontsize{9}{9}\selectfont 34328 & \fontsize{9}{9}\selectfont 10.80 & \fontsize{9}{9}\selectfont 12.00 (\textcolor{blue}{+11.11\%}) & \fontsize{9}{9}\selectfont 8.80 (\textcolor{red}{-18.51\%}) & \fontsize{9}{9}\selectfont 18.40 \\
		\fontsize{9}{9}\selectfont Word sorting  & \fontsize{9}{9}\selectfont 7775 & \fontsize{9}{9}\selectfont 3.20 & \fontsize{9}{9}\selectfont 1.60 (\textcolor{red}{-50.00\%}) & \fontsize{9}{9}\selectfont 5.20 (\textcolor{blue}{+62.50\%}) & \fontsize{9}{9}\selectfont 24.40  \\
\hline		
    \fontsize{9}{9}\selectfont Disambiguation QA  & \fontsize{9}{9}\selectfont 12049 & \fontsize{9}{9}\selectfont 30.8  & \fontsize{9}{9}\selectfont 38.80 (\textcolor{blue}{+25.97\%}) & \fontsize{9}{9}\selectfont 42.00 (\textcolor{blue}{+36.36\%}) & \fontsize{9}{9}\selectfont 47.20 \\
		\fontsize{9}{9}\selectfont Hyperbaton  & \fontsize{9}{9}\selectfont 5135 & \fontsize{9}{9}\selectfont 50.00 & \fontsize{9}{9}\selectfont 49.60 (\textcolor{red}{-0.80\%}) & \fontsize{9}{9}\selectfont 54.80 (\textcolor{blue}{+9.60\%}) & \fontsize{9}{9}\selectfont 57.60 \\
		\fontsize{9}{9}\selectfont Snarks  & \fontsize{9}{9}\selectfont 6206 & \fontsize{9}{9}\selectfont 52.24 & \fontsize{9}{9}\selectfont 43.25 (\textcolor{red}{-17.20\%}) & \fontsize{9}{9}\selectfont 53.93 (\textcolor{blue}{+3.23\%}) & \fontsize{9}{9}\selectfont 70.78  \\
\hline		
    \fontsize{9}{9}\selectfont Causal judgement  & \fontsize{9}{9}\selectfont 35034 & \fontsize{9}{9}\selectfont 54.01  & \fontsize{9}{9}\selectfont 31.55 (\textcolor{red}{-41.58\%}) & \fontsize{9}{9}\selectfont 47.59 (\textcolor{red}{-11.88\%}) & \fontsize{9}{9}\selectfont 59.89 \\
    \fontsize{9}{9}\selectfont Date understanding  & \fontsize{9}{9}\selectfont 7530 & \fontsize{9}{9}\selectfont 16.00  & \fontsize{9}{9}\selectfont 20.80 (\textcolor{blue}{+30.00\%}) & \fontsize{9}{9}\selectfont 22.40 (\textcolor{blue}{+40.00\%}) & \fontsize{9}{9}\selectfont 68.00 \\
		\fontsize{9}{9}\selectfont Movie recommendation  & \fontsize{9}{9}\selectfont 7881 & \fontsize{9}{9}\selectfont 28.80 & \fontsize{9}{9}\selectfont 14.40 (\textcolor{red}{-50.00\%}) & \fontsize{9}{9}\selectfont 29.20 (\textcolor{blue}{+1.38\%}) & \fontsize{9}{9}\selectfont 64.80  \\
		\fontsize{9}{9}\selectfont Object counting  & \fontsize{9}{9}\selectfont 6425 & \fontsize{9}{9}\selectfont 22.40 & \fontsize{9}{9}\selectfont 24.80 (\textcolor{blue}{+10.71\%}) & \fontsize{9}{9}\selectfont 25.60 (\textcolor{blue}{+14.28\%}) & \fontsize{9}{9}\selectfont 55.60 \\
		\fontsize{9}{9}\selectfont Penguins in a table  & \fontsize{9}{9}\selectfont 13396 & \fontsize{9}{9}\selectfont 21.23 & \fontsize{9}{9}\selectfont 17.80 (\textcolor{red}{-16.15\%}) & \fontsize{9}{9}\selectfont 27.39 (\textcolor{blue}{+29.01\%}) & \fontsize{9}{9}\selectfont 51.36 \\								
		\fontsize{9}{9}\selectfont Reasoning about colored objects  & \fontsize{9}{9}\selectfont 13554 & \fontsize{9}{9}\selectfont 15.60 & \fontsize{9}{9}\selectfont 16.80 (\textcolor{blue}{+7.69\%}) & \fontsize{9}{9}\selectfont 15.60 (\textcolor{black}{0\%}) & \fontsize{9}{9}\selectfont 62.40 \\																															
		\fontsize{9}{9}\selectfont Ruin names  & \fontsize{9}{9}\selectfont 7617 & \fontsize{9}{9}\selectfont 21.20 & \fontsize{9}{9}\selectfont 22.00 (\textcolor{blue}{+3.77\%}) & \fontsize{9}{9}\selectfont 20.80 (\textcolor{red}{-1.88\%}) & \fontsize{9}{9}\selectfont 41.20 \\	
	\fontsize{9}{9}\selectfont Sport understanding  & \fontsize{9}{9}\selectfont 3780 & \fontsize{9}{9}\selectfont 60.40 & \fontsize{9}{9}\selectfont 54.00 (\textcolor{red}{-10.59\%}) & \fontsize{9}{9}\selectfont 55.20 (\textcolor{red}{-8.60\%}) & \fontsize{9}{9}\selectfont 92.00 \\		
		\fontsize{9}{9}\selectfont Web of lies  & \fontsize{9}{9}\selectfont 8035 & \fontsize{9}{9}\selectfont 51.60 & \fontsize{9}{9}\selectfont 51.60 (\textcolor{black}{0\%}) & \fontsize{9}{9}\selectfont 51.60 (\textcolor{black}{0\%}) & \fontsize{9}{9}\selectfont 94.40 \\			
\hline		
		\fontsize{9}{9}\selectfont Salient translation error detection  & \fontsize{9}{9}\selectfont 38692 & \fontsize{9}{9}\selectfont 12.80 & \fontsize{9}{9}\selectfont 4.40 (\textcolor{red}{-65.62\%}) & \fontsize{9}{9}\selectfont 12.00 (\textcolor{red}{-6.25\%}) & \fontsize{9}{9}\selectfont 39.60  \\
    \hline
		\fontsize{9}{9}\selectfont Average  & \fontsize{9}{9}\selectfont 15300 & \fontsize{9}{9}\selectfont 27.96\% & \fontsize{9}{9}\selectfont 26.48\% (\textcolor{red}{-5.29\%}) & \fontsize{9}{9}\selectfont \textbf{29.23\%} (\textbf{\textcolor{blue}{+4.54\%}}) & \fontsize{9}{9}\selectfont 49.95\% \\
		\hline
  \end{tabular}
  \caption{\label{performance-tinyllama-tasks}
    Exact-match scores (in \%) of the student (baseline), distilled (KD), and teacher models on the 27 tasks of the BBH benchmark, with TinyLlama as the student model and Llama2-13B-Chat as the teacher model. The numbers in parentheses represent relative gains, except where absolute gains (abs.) are specified. \vspace{-10pt}
  }
\end{table*}

Results of the Qwen-based KD experiments are shown in Table \ref{performance-qwen-tasks} while those of the Llama2-based KD experiments are shown in Tables \ref{performance-llama2-tasks} and \ref{performance-tinyllama-tasks}. In the Qwen-based experiments, both the distilled models created with the vanilla white-box KD (Qwen-1.8B+KD) and the KD+CoT methods (Qwen-1.8B+KD+CoT) yield improved average exact-match scores compared to the student model (Qwen-1.8B). While the vanilla white-box KD improves the average performance across 27 tasks by 30\% relative, the KD+CoT further improves the result, with 7.54\% relative boost.

In the Llama2-based experiments (see Tables \ref{performance-llama2-tasks} and \ref{performance-tinyllama-tasks}), the vanilla white-box KD yields improved performance for some tasks but not on average across all the tasks. In contrast, the KD+CoT method improves average performance by 5.22\% relative with Llama2-7B and 4.54\% relative with TinyLlama as student models, respectively.

While improvements are not universal, the KD+CoT approach is observed to enhance performance across several natural language reasoning and understanding tasks. This highlights the role of CoT in white-box KD, and we will analyse these tasks in detail in the following sections.

\subsubsection{Analysis}
\label{app:discussion}

We analyse the performance of the models on the individual tasks of BBH across its four categories:

\paragraph{Algorithmic and multi-step arithmetic reasoning}
Many of the tasks in BBH require varying levels of arithmetic (e.g., \textit{multistep arithmetic two}), logical (e.g., \textit{boolean expression} and \textit{logical deduction}), geometric (e.g., \textit{geometric shapes}), hierarchical (e.g., \textit{dyck language}), spatial (e.g., \textit{navigate}), and temporal (e.g., \textit{temporal sequences}) reasoning capabilities \cite{suzgun2023_acl_findings}. The general observation suggests that KD with CoT improves performance in some tasks within this category, but not all. When the vanilla white-box KD already improves upon the baseline, CoT further amplifies its effectiveness. Conversely, when the vanilla white-box KD does not outperform the baseline, CoT helps elevate its performance, making the distillation process more effective. This observation is noticeable but not universally applicable to both Qwen-based and Llama2-based KD experiments.

An example of the \textit{Temporal sequences} task which requires LLM's temporal reasoning capability is shown in \textbf{Example (i)}. In this example, we can see that the Qwen-1.8B+KD+CoT model's answer has deeper level of reasoning compared to those of the Qwen-1.8B and Qwen-1.8B+KD models. Thanks to this improved reasoning capability, the Qwen-1.8B+KD+CoT model answers correctly to the question. It should be noted that there are tasks in this category where the vanilla KD and KD+CoT do not improve over the student model, for example for \textit{Boolean expressions} task when the student model is Qwen-1.8B.
\\[0.2cm]
\noindent \fbox{\begin{minipage}{19.5em}
\fontsize{10}{11}\selectfont \ttfamily\slshape \underline{Example (i) (\textbf{Temporal sequences} task)}\\
\fontsize{10}{11}\selectfont \ttfamily\slshape \textbf{Question:} Today, Sarah went to the clothing store. Between what times could they have gone? We know that : Sarah woke up at 5am. Ashley saw Sarah taking photos near the Eiffel Tower from 5am to 11am. Michael saw Sarah walking in the garden from 11am to 12pm. Emily saw Sarah buying cookies at a bakery from 12pm to 2pm. Richard saw Sarah buying a phone at the electronics store from 2pm to 3pm. The clothing store was closed after 7pm. Between what time could Sarah have gone to the clothing store? \textbf{Options:} \textbf{(A)} 3pm to 7pm, \textbf{(B)} 2pm to 3pm, \textbf{(C)} 12pm to 2pm, \textbf{(D)} 11am to 12pm.
\end{minipage}}
\\[0.3cm]
\noindent \fbox{\begin{minipage}{19.5em}
\fontsize{10}{11}\selectfont \ttfamily\slshape \underline{Example (i) (\textbf{Temporal sequences} task)}\\
\fontsize{10}{11}\selectfont \ttfamily\slshape \textbf{Answer (Qwen-1.8B):} Let's think step by step. Wake-up time: 5am. 5am-11am: taking photos near the Eiffel Tower. 11am-12pm: walking in the garden. 12pm-2pm: buying cookies at the bakery. 2pm-3pm: buying a phone at the electronics store. 3pm-7pm: the clothing store is closed. The clothing store closure time: 7pm. The only time when Sarah could have gone to the clothing store was 2pm to 3pm. \textbf{So the answer is (C) 12pm to 2pm}.
\end{minipage}}
\\[0.3cm]
\noindent \fbox{\begin{minipage}{19.5em}
\fontsize{10}{11}\selectfont \ttfamily\slshape \underline{Example (i) (\textbf{Temporal sequences} task)}\\
\fontsize{10}{11}\selectfont \ttfamily\slshape \textbf{Answer (Qwen-1.8B+KD):} Let's think step by step. Wake-up time: 5am. 5am-11am: taking photos near the Eiffel Tower. 11am-12pm: walking in the garden. 12pm-2pm: buying cookies at the bakery. 2pm-3pm: buying a phone at the electronics store. The clothing store closure time: 7pm. The only time when Sarah could have gone to the clothing store was 11am to 12pm. \textbf{So the answer is (C)}.
\end{minipage}}
\\[0.3cm]
\noindent \fbox{\begin{minipage}{19.5em}
\fontsize{10}{11}\selectfont \ttfamily\slshape \underline{Example (i) (\textbf{Temporal sequences} task)}\\
\fontsize{10}{11}\selectfont \ttfamily\slshape \textbf{Answer (Qwen-1.8B+KD+CoT):} Let's think step by step. Wake-up time: 5am. 5am-11am: taking photos near the Eiffel Tower. 11am-12pm: walking in the garden. 12pm-2pm: buying cookies at the bakery. 2pm-3pm: buying a phone at the electronics store. The clothing store closure time: 7pm. The only time when Sarah could have gone to the clothing store was 3pm to 7pm. \textbf{So the answer is (A)}.
\end{minipage}}
%\\[0.1cm]
\vspace{-10pt}
\paragraph{Natural language understanding} BBH encompasses several tasks that emphasise semantic understanding, name disambiguation, entity resolution, grammar rules, and sarcasm detection. Examples of these tasks include \textit{disambiguation QA}, \textit{hyperbaton} (adjective ordering), \textit{salient translation error detection}, and \textit{snarks} (sarcasm detection). In tasks such as \textit{Disambiguation QA}, stronger reasoning capability enables LLMs to interpret nuanced spoken language questions and identify ambiguity. \textbf{Example (ii)} illustrates a \textit{Disambiguation QA} task instance, where the model must determine whether a pronoun’s antecedent is ambiguous in a specific context. In this example, the Llama2-7B+KD model does not improve upon the Llama2-7B model’s answer, whereas the Llama2-7B+KD+CoT effectively analyses the question's nuances and provides the correct answer. This demonstrates the improved reasoning capability in the model created with KD+CoT approach. 

\vspace{5pt} While performance gains are typically observed across this task group when using Llama2-7B and TinyLlama-1.1B as student models, such improvements are notably absent with Qwen-1.8B on the \textit{Disambiguation QA} task. This highlights the variability in effectiveness of CoT in white-box KD across different model architectures.
\\[0.3cm]
\noindent \fbox{\begin{minipage}{19.5em}
\fontsize{10}{11}\selectfont \ttfamily\slshape \underline{Example (ii) (\textbf{Disambiguation QA} task)}\\
\fontsize{10}{11}\selectfont \ttfamily\slshape \textbf{Question:} In the following sentences, explain the antecedent of the pronoun (which thing the pronoun refers to), or state that it is ambiguous. \textbf{Sentence:} The guard called the cleaner and asked them to open the door. \textbf{Options:} \textbf{(A)} Asked the guard \textbf{(B)} Asked the cleaner \textbf{(C)} Ambiguous.
\end{minipage}}
%\\[0.3cm]
\paragraph{Use of world knowledge} Certain BBH tasks necessitate both factual and general knowledge about the world, along with an understanding of common practices and assumptions in Western society. For instance, factual knowledge tasks include \textit{sport understanding}, \textit{movie recommendation}, and \textit{date understanding}. On the other hand, tasks that demand general knowledge include \textit{causal judgement}, which relies on an understanding of causal reasoning, and \textit{ruin names}, which requires understanding human 
\noindent \fbox{\begin{minipage}{19.5em}
\fontsize{10}{11}\selectfont \ttfamily\slshape \underline{Example (ii) (\textbf{Disambiguation QA} task)}\\
\fontsize{10}{11}\selectfont \ttfamily\slshape \textbf{Answer (Llama2-7B} and \textbf{Llama2-7B+KD):} Let's think step by step. Here we need to determine who the pronoun ``them'' might be referring to. There are two possible referents for ``them'', namely the guard and the cleaner. The verb ``asked'' might be able to help us determine which one is more likely (if either). Let X be the guard and Y the cleaner. The sentence is then of the form ``X asked Y to do something''. Let's consider Y first: ``X asked Y to do something''. This case makes sense, because X is the one who asked Y to do something. Now, consider X: ``X asked Y to do something''. This case also makes sense, because X is the one who asked Y to do something. Because both X and Y are possible at the same time, we conclude that the antecedent of the pronoun is ambiguous. \textbf{So the answer is (C)}.
\end{minipage}}
\\[0.3cm]
\noindent \fbox{\begin{minipage}{19.5em}
\fontsize{10}{11}\selectfont \ttfamily\slshape \underline{Example (ii) (\textbf{Disambiguation QA} task)}\\
\fontsize{10}{11}\selectfont \ttfamily\slshape \textbf{Answer (Llama2-7B+KD+CoT):} Let's think step by step. Here we need to determine who the pronoun ``them'' might be referring to. There are two possible referents for ``them'', namely the guard and the cleaner. The verb ``asked'' might be able to help us determine which one is more likely (if either). Let X be the guard and Y the cleaner. The sentence is then of the form ``X asked Y to do something''. Let's consider Y first: ``X asked Y to do something''. This case makes sense, because X is the one who asked Y to do something. Now, consider X: ``X asked them to do something''. This case doesn't make sense, because X is the one who asked Y to do something, so X cannot be the one who is being asked to do something. Because in this context, X is the guard and Y is the cleaner, the answer should be the cleaner. \textbf{So the answer is (B)}.
\end{minipage}}
\\[0.3cm]
perception, humor in the English language, and familiarity with Western artists, bands, and movies. For these tasks, a stronger grounding in rich common knowledge enables LLMs to better meet requirements. The KD+CoT approach yields improvements for a number of tasks in this group, for examples \textit{Date understanding} or \textit{Penguins in a table}. However, the approach is not effective in some tasks, for examples \textit{Causal Judgement} when the student model is Llama2-7B or \textit{Sport understanding} when the student model is TinyLlama-1.1B. 

In \textbf{Example (iii)} from the Sport Understanding task, both TinyLlama and TinyLlama+KD produce the correct answer, whereas TinyLlama+KD+CoT fails to do so. In this case, all models are required to identify the name of a baseball player and then 
\noindent \fbox{\begin{minipage}{19.5em}
\fontsize{10}{11}\selectfont \ttfamily\slshape \underline{Example (iii) (\textbf{Sport understanding} task)}\\
\fontsize{10}{11}\selectfont \ttfamily\slshape \textbf{Question:} Is the following sentence plausible? ``Marcell Ozuna scored a touchdown''  \textbf{Options:} \textbf{(A)} Yes \textbf{(B)} No.
\end{minipage}}
\\[0.3cm]
\noindent \fbox{\begin{minipage}{19.5em}
\fontsize{10}{11}\selectfont \ttfamily\slshape \underline{Example (iii) (\textbf{Sport understanding} task)}\\
\fontsize{10}{11}\selectfont \ttfamily\slshape \textbf{Answer (TinyLlama:} Let's think step by step. Marcell Ozuna is a baseball player. Touchdown is part of American football. \textbf{So the answer is (B) No}.
\end{minipage}}
\\[0.3cm]
\noindent \fbox{\begin{minipage}{19.5em}
\fontsize{10}{11}\selectfont \ttfamily\slshape \underline{Example (iii) (\textbf{Sport understanding} task)}\\
\fontsize{10}{11}\selectfont \ttfamily\slshape \textbf{Answer (TinyLlama+KD:} Let's think step by step. Marcell Ozuna is a baseball player. Touchdown is part of American football and rugby. \textbf{So the answer is (B) No}.
\end{minipage}}
\\[0.3cm]
\noindent \fbox{\begin{minipage}{19.5em}
\fontsize{10}{11}\selectfont \ttfamily\slshape \underline{Example (iii) (\textbf{Sport understanding} task)}\\
\fontsize{10}{11}\selectfont \ttfamily\slshape \textbf{Answer (TinyLlama+KD+CoT):} Let's think step by step. Marcell Ozuna is a baseball player. Touchdown is part of American football and baseball. \textbf{So the answer is (A) Yes}.
\end{minipage}}
\\[0.3cm]
ground their response in relevant sports knowledge. The error made by TinyLlama+KD+CoT appears to stem from retrieving knowledge that aligns with the player's discipline but ultimately misguides the model toward an incorrect conclusion.
\paragraph{Multilingual knowledge and reasoning} General-purpose language models possess the capability to perform translation and engage in multistep reasoning within multilingual contexts \cite{winata2021_wmrl, shi2023_iclr}. In BBH, there is one multilingual task, \textit{salient translation error detection}, which is based on translation quality estimation and cross-lingual natural-language inference. In this task, improved reasoning capability could enhance the ability to navigate multiple languages and generate accurate translations. An example on this task where the Llama2-7B+KD+CoT model successfully produces the correct answer, whereas the Llama2-7B and Llama2-7B+KD models do not, is presented in \textbf{Example (iv)}. In this example, the LLMs are tasked with identifying an error in a German-to-English translation. The Llama2-7B+KD+CoT model produces a correct answer with improved reasoning, while the Llama2-7B+KD is unable to improve upon the Llama2-7B's response.

More specifically, the Llama2-7B+KD+CoT model successfully retrieves the correct German translation of the input sentence via Google Translate. It is then able to distinguish between the correct and incorrect translations, and accurately identify the error as related to named entities. Indeed,

\noindent \fbox{\begin{minipage}{19.5em}
\fontsize{10}{11}\selectfont \ttfamily\slshape \uline{Example (iv) (\textbf{Salient} \textbf{translation} \textbf{error} \textbf{detection} task)} \\
\fontsize{10}{11}\selectfont \ttfamily\slshape \textbf{Question:} The following translations from German to English contain a particular error. That error will be one of the following types: \textbf{Named Entities:} An entity (names, places, locations, etc.) is changed to a different entity. \textbf{Numerical Values:} Numerical values (ordinals or cardinals), dates, and/or units are changed. \textbf{Modifiers or Adjectives:} The modifiers and adjectives pertaining to a noun are changed. \textbf{Negation or Antonyms:} Introduce or remove a negation or change comparatives to their antonyms. \textbf{Facts:} Trivial factual errors not pertaining to the above classes are introduced in the translations. \textbf{Dropped Content:} A significant clause in the translation is removed. Please identify that error.  Source: Elbenberg ist ein Stadtteil der nordhessischen Kleinstadt Naumburg im Landkreis Kassel. Translation: Elbenberg is a district of Naumburg in the district of Tossel in northern Hesse, Germany. The translation contains an error pertaining to \textbf{Options:} \textbf{(A)} Modifiers or Adjectives \textbf{(B)} Numerical Values \textbf{(C)} Negation or Antonyms \textbf{(D)} Named Entities \textbf{(E)} Dropped Content \textbf{(F)} Facts.
\end{minipage}}
\\[0.3cm]
\noindent \fbox{\begin{minipage}{19.5em}
\fontsize{10}{11}\selectfont \ttfamily\slshape \uline{Example (iv) (\textbf{Salient} \textbf{translation} \textbf{error} \textbf{detection} task)} \\
\fontsize{10}{11}\selectfont \ttfamily\slshape \textbf{Answer (Llama2-7B} and \textbf{Llama2-7B+KD):} Let's think step by step. We solve this question by first translating the source sentence to English and then by comparing our translation with the provided translation. According to Google Translate, the correct translation of the source sentence from German to English is ``Elbenberg is a district of Naumburg in the district of Tossel in northern Hesse, Germany.'' On the other hand, the provided translation is ``Elbenberg is a district of Naumburg in the district of Tossel in northern Hesse, Germany.'' Note that the provided sentence says ``Elbenberg is a district of Naumburg in the district of Tossel in northern Hesse, Germany.'' However, the translation should have been ``Elbenberg is a district of Naumburg in the district of Tossel in northern Hesse, Germany.'' Because a negation is introduced at the beginning of the sentence and has fundamentally changed the meaning of the original source, the translation contains an error pertaining to Negation or Antonyms. \textbf{So the answer is (C)}.
\end{minipage}}
\\[0.3cm]
this capability to recognise the correct translation is supported by its enhanced reasoning capabilities.
%\vspace{-10pt}
\subsubsection{Inference time}
\label{sec:inference_time}

The distilled models have similar inference time as the student models since they share the same num-

\noindent \fbox{\begin{minipage}{19.5em}
\fontsize{10}{11}\selectfont \ttfamily\slshape \uline{Example (iv) (\textbf{Salient} \textbf{translation} \textbf{error} \textbf{detection} task)} \\
\fontsize{10}{11}\selectfont \ttfamily\slshape \textbf{Answer (Llama2-7B+KD+CoT):} Let's think step by step. We solve this question by first translating the source sentence to English and then by comparing our translation with the provided translation. According to Google Translate, the correct translation of the source sentence from German to English is ``Elbenberg is a district of Naumburg in the district of Kassel in northern Hesse, Germany.'' On the other hand, the provided translation is ``Elbenberg is a district of Naumburg in the district of Tossel in northern Hesse, Germany.'' Note that the provided sentence says ``Elbenberg is a district of Naumburg in the district of Tossel in northern Hesse, Germany.'' However, the translation should have been ``Elbenberg is a district of Naumburg in the district of Kassel in northern Hesse, Germany.'' Because a district in the original source is changed to a different district in the translation, the translation contains an error pertaining to Named Entities. \textbf{So the answer is (D)}.
\end{minipage}}
\\[0.2cm]
\noindent ber of parameters. However, they exhibit significantly lower inference time compared to the teacher models. More specifically, the TinyLlama model requires only 20\% while the Llama2-7B requires approximately 53\% of the Llama2-13B-Chat’s inference time, measured on an NVIDIA A100 80GB GPU. Similarly, the Qwen-1.8B model operates at approximately 28\% of the inference time of Qwen-7B, with the same inference settings. The improved inference speed makes the distilled models more suitable for deployment in resource-constrained environments compared to the larger teacher models.

\vspace{-7pt}
\section{Conclusion}
\label{sec:conclusions}
\vspace{-8pt}

This paper examined the effectiveness of CoT in distilling reasoning capability from a teacher LLM to a student LLM using white-box KD, where the distillation loss between their output probabilities is minimised. Experimental results demonstrated that CoT improves the effectiveness of white-box KD, even when the vanilla white-box KD does not outperform the baseline. The results also suggest that CoT elicits the teacher LLM’s reasoning capability, which are then effectively distilled into the student model during the distillation process. In summary, the KD+CoT approach enhances the distilled models, improving their overall performance in BBH natural language reasoning and understanding tasks over the student models while maintaining their efficiency advantage over the teacher models in terms of inference speed.

\section{Data availability statement}
\label{sec:data_availability_statement}

In this paper, we have used the CoT-Collection dataset \cite{kim2023_emnlp} for training and the data from various natural language reasoning and understanding tasks in the BBH benchmark \cite{suzgun2023_acl_findings} for evaluation. From the CoT-Collection data, we have created new training data to train the vanilla white-box KD by filtering out the rationales. In terms of codes, we have developed our white-box KD training scripts based on the MiniLLM framework \cite{gu2024_iclr} and our evaluation scripts based on \cite{suzgun2023_acl_findings}. All datasets and code are included in the supplementary materials accompanying this paper and are publicly accessible.

%\begin{itemize}
%\item Data for training the vanilla KD
%\item Data for training the KD+CoT
%\item Training and evaluation scripts
%\end{itemize}

%\section{Ethical considerations}
%\label{sec:ethical_considerations}

\bibliography{custom}

%\newpage 
\vspace{0.5cm}
\appendix

\section{BIG-Bench Hard task descriptions}
%\vspace{1cm}
\noindent Brief descriptions of the tasks in the BIG-Bench-Hard (BBH) benchmark, as presented in \cite{suzgun2023_acl_findings}:

\begin{itemize}
\item \textbf{Boolean Expressions:} Evaluate the truth value of a random Boolean expression consisting of Boolean constants (`True`, `False`) and basic Boolean operators (`and`, `or`, `not`).

\item \textbf{Causal Judgment:} Given a short story (involving moral, intentional, or counterfactual analysis), determine how a typical person would answer a causal question about the story.

\item \textbf{Date Understanding:} Given a small set of sentences about a particular date, answer the provided question.

\item \textbf{Disambiguation QA:} Given a sentence with an ambigious pronoun, either determine whether the sentence is inherently ambiguous (i.e., the thing that the pronoun refers to cannot be inferred by given information) or, if the pronoun can be implicitly deduced, state the antecedent of the pronoun (i.e., the noun to which the pronoun refers).

\item \textbf{Dyck Languages:} Predict the sequence of the closing parentheses of a Dyck-4 word without its last few closing parentheses. 

\item \textbf{Formal Fallacies Syllogisms Negation:} Given a context involving a set of statements (generated by one of the argument schemes), determine whether an argument---presented informally---can be logically deduced from the provided context.\footnote{This task has a particular focus on negations and was designed to understand whether LLMs can distinguish deductively valid arguments from formal fallacies.}

\item \textbf{Geometric Shapes:} Given a full SVG path element containing multiple commands, determine the geometric shape that would be generated if one were to execute the full path element. 

\item \textbf{Hyperbaton (Adjective Ordering):} Given two English-language sentences, determine the one with the correct adjective order. 

\item \textbf{Logical Deduction:} Deduce the order of a sequence of objects based on the clues and information about their spacial relationships and placements.

\item \textbf{Movie Recommendation:} Given a list of movies a user might have watched and liked, recommend a new, relevant movie to the user out of the four potential choices user might have.

\item \textbf{Multi-Step Arithmetic:} Solve multi-step equations involving basic arithmetic operations (addition, subtraction, multiplication, and division). 

\item \textbf{Navigate:} Given a series of navigation steps to an agent, determine whether the agent would end up back at its initial starting point.

\item \textbf{Object Counting:} Given a collection of possessions that a person has along with their quantities (e.g., three pianos, two strawberries, one table, and two watermelons), determine the number of a certain object/item class (e.g., fruits).

\item \textbf{Penguins in a Table:} Given a unique table of penguins (and sometimes some new information), answer a question about the attributes of the penguins. 

\item \textbf{Reasoning about Colored Objects:} Given a context, answer a simple question about the color of an object on a surface.

\item \textbf{Ruin Names:} Given an artist, band, or movie name, identify a one-character edit to the name that changes the meaning of the input and makes it humorous. 

\item \textbf{Salient Translation Error Detection:} Given a source sentence written in German and its translation in English, determine the type of translation error that the translated sentence contains.

\item \textbf{Snarks:} Given two nearly-identical sentences, determine which one is sarcastic.

\item \textbf{Sports Understanding:} Determine whether a factitious sentence related to sports is plausible.

\item \textbf{Temporal Sequences:} Given a series of events and activities a person has completed in the course of a day, determine what time, during the day, they might have been free to perform another activity.

\item \textbf{Tracking Shuffled Objects:} Given the initial positions of a set of objects and a series of transformations (namely, pairwise swaps) applied to them, determine the final positions of the objects.

\item \textbf{Web of Lies:} Evaluate the truth value of a random Boolean function expressed as a natural-language word problem.

\item \textbf{Word Sorting:} Given a list of words, sort them lexicographically.
\end{itemize}

\end{document}